\documentclass[letterpaper, 10pt, journal]{IEEEtran}

\usepackage[utf8]{inputenc}
\usepackage[T1]{fontenc}
\usepackage{adjustbox}
\usepackage{amsmath}
\usepackage{amsfonts}
\usepackage{amssymb}
\usepackage{graphicx}
\usepackage{siunitx}
\usepackage{color}
\usepackage{xcolor}
\usepackage{booktabs}
\usepackage{dblfloatfix}
\usepackage{url}
\usepackage{nth}
\usepackage{hyperref}
\hypersetup{
    colorlinks=true,
    linkcolor=black,  
    citecolor=black,
    urlcolor=black,
}

\usepackage{subfig}
\usepackage{hyphenat}
\usepackage{cite}

\title{Expertise Affects Drone Racing Performance}

\author{Christian Pfeiffer and Davide Scaramuzza
\thanks{The authors are with the Robotics and Perception Group, Department of Informatics, University of Zurich, and Department of Neuroinformatics, University of Zurich and ETH Zurich, Switzerland (\protect\url{http://rpg.ifi.uzh.ch}). This work was supported by the Ernst Göhner Foundation and University of Zurich Alumni Fonds zur Förderung des Akademischen Nachwuchses (FAN Fellowship), by  the National Centre of Competence in Research (NCCR) Robotics through the Swiss National Science Foundation (SNSF) and the European Union’s Horizon 2020 Research and Innovation Programme under grant agreement No. 871479 (AERIAL-CORE) and the European Research Council (ERC) under grant agreement No. 864042 (AGILEFLIGHT).}
}

\begin{document}

\maketitle

\begin{abstract}
First-person view drone racing has become a popular televised sport. However, very little is known about the perceptual and motor skills of professional drone racing pilots. A better understanding of these skills may inform path planning and control algorithms for autonomous multirotor flight. By using a real-world drone racing track and a large-scale position tracking system, we compare the drone racing performance of five professional and five beginner pilots. Results show that professional pilots consistently outperform beginner pilots by achieving faster lap times, higher velocity, and more efficiently executing the challenging maneuvers. Trajectory analysis shows that experienced pilots choose more optimal racing lines than beginner pilots. Our results provide strong evidence for a contribution of expertise to performances in real-world human-piloted drone racing. We discuss the implications of these results for future work on autonomous fast and agile flight. We make our data openly available.
\end{abstract}


The dataset can be downloaded at \protect\url{https://osf.io/uabx4/}.

\section{Introduction}
First-Person View (FPV) drone racing has become a popular televised sport with international competitions and up to a hundred thousand USD prize money for the winning pilot\footnote{\url{https://thedroneracingleague.com}}. Human pilots view a first-person video stream from a camera attached to the drone and send collective thrust and body rate commands with a hand-held remote. The goal of a drone race is to complete a given sequence of waypoints---indicated by racing gates or flags---in a minimum time without crashing the drone. Professional pilots spend years refining the perceptual and manual skills required to achieve top performances in drone racing.

\begin{figure}[t]
  \centering
    \subfloat{\includegraphics[width=\linewidth,trim={0 90 0 0},clip]{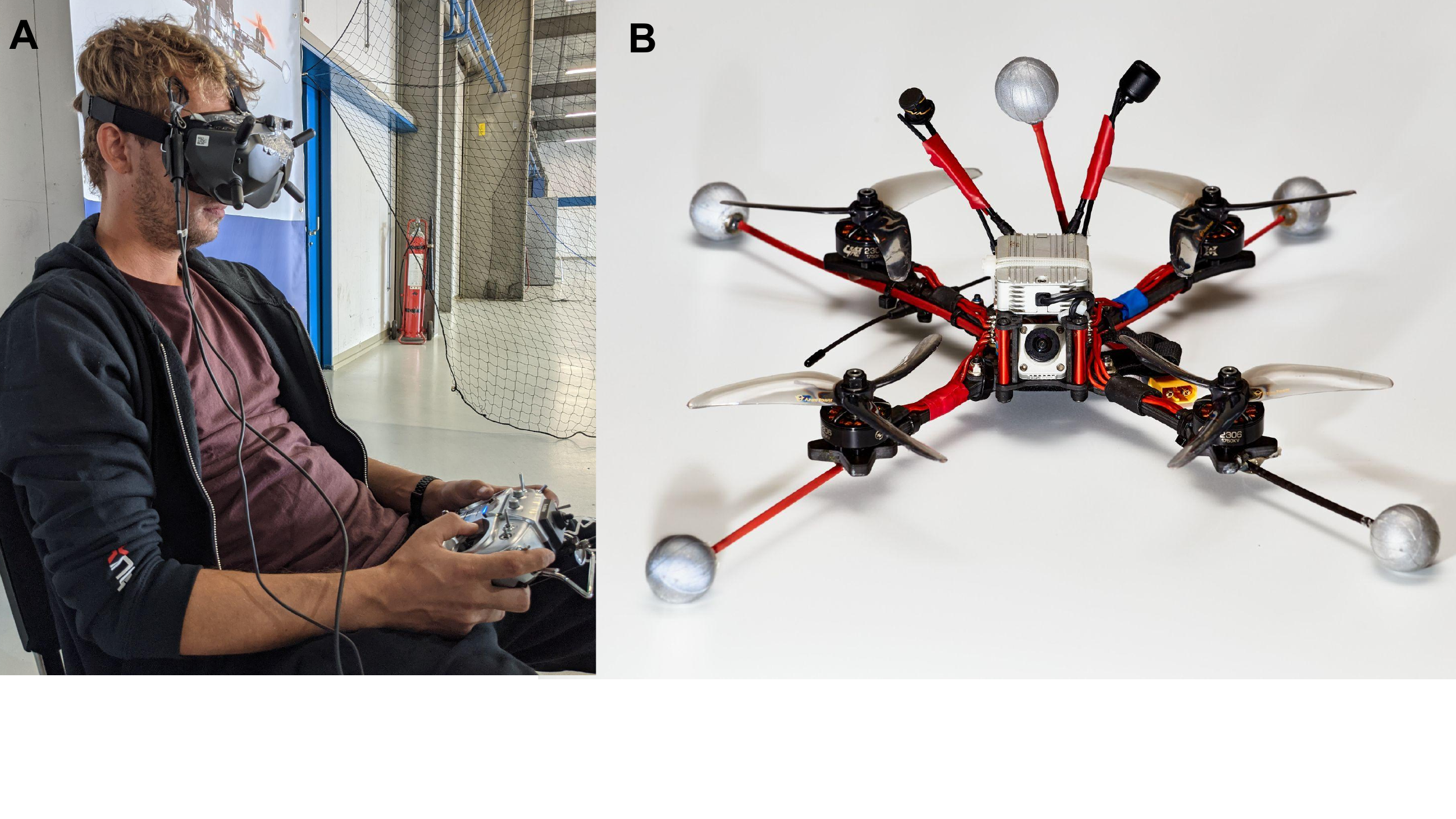}}
    \newline
    \subfloat{\includegraphics[width=\linewidth,trim={0 120 0 0},clip]{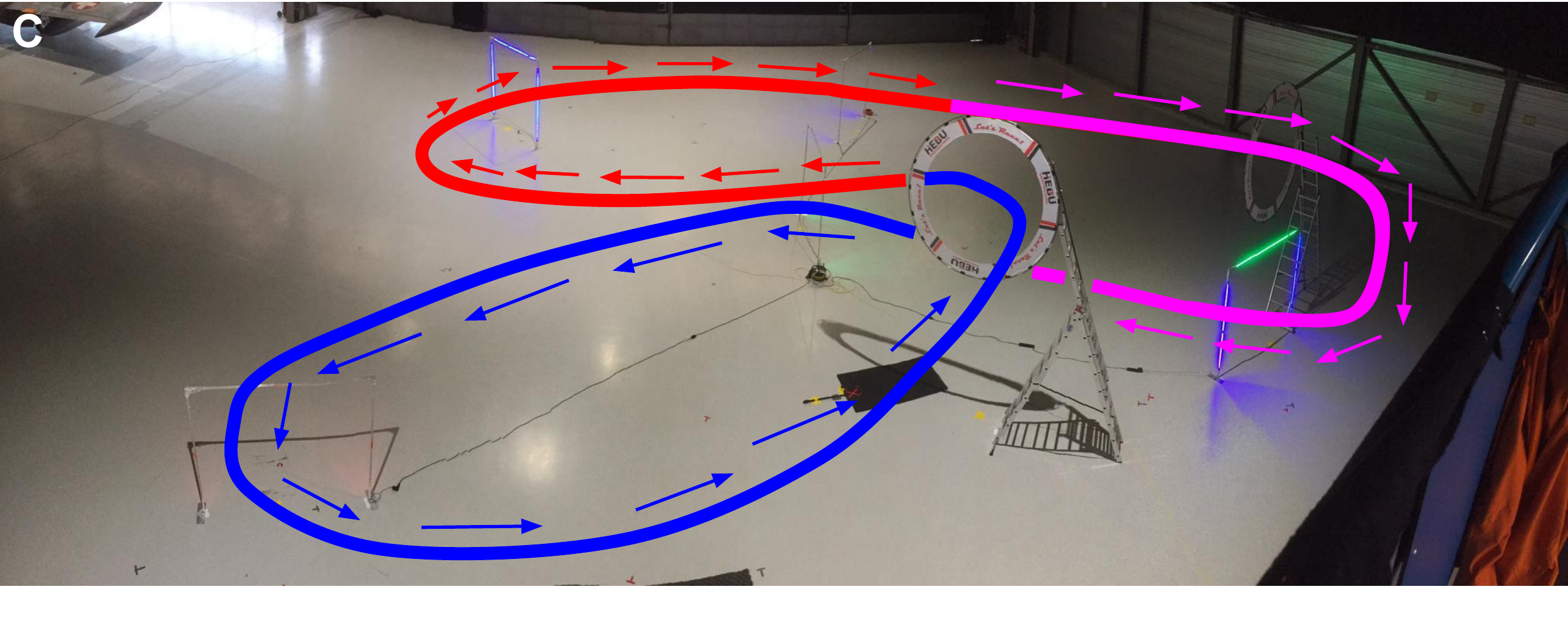}}
\caption[Short caption.] {\label{fig:pilot_drone_track} \textbf{A:} Human pilot equipped with a first-person headset and remote controller. \textbf{B:} Racing drone with position-tracking markers. \textbf{C:} Flying arena showing the race track and schematic flight path. Arrows indicate direction of flight and different colors indicate the left turn (blue), right turn (red) and Split-S maneuver (purple).}
\end{figure}

These human performances did not go unnoticed by the research community. Over the past decade, research on autonomous multirotor drones has achieved increasingly higher speeds and agility \cite{Mellinger2010,Loianno2017,Kaufmann2018a,Foehn2020,Romero2021}. Competitions have been organized, such as the Autonomous Drone Racing series at the recent IROS and NeurIPS conferences \cite{Moon2019,Cocoma-Ortega2019,Madaan2020a} and the AlphaPilot challenge\cite{Foehn2020,Guerra2019} intending to develop autonomous systems that will eventually outperform expert human pilots. Million-dollar projects, such as AgileFlight \cite{cordisAgileFlightEUproject} and Fast Light Autonomy (FLA) \cite{Mohta2018}, have also been funded by the European Research Council and the United States government, respectively, to further advance the field.

Given this ambitious goal, surprisingly little research is available that investigates the perceptual and motor abilities of top-tier professional drone racing pilots \cite{Pfeiffer2021b,Barin2017}. This knowledge can serve as a benchmark for autonomous racing drones and help improve the performance of path planning and control algorithms, particularly for vision-based, autonomous, fast, and agile flight. 

To better understand the perceptual and motor skills of professional drone racing pilots, this study collected data from five top-tier professional pilots and five beginner pilots in a real-world drone racing task using a large-scale position tracking system. 

This study addresses the following research questions:
1. \emph{How does flight performance differ between professional and beginner pilots?}
2. \emph{How does control behavior differ between professional and beginner pilots?}
3. \emph{How do flight trajectories of professional and beginner pilots compare to time-optimal trajectories of an autonomous drone?}

\begin{figure*}[t]
    \centering
    \includegraphics[width=\textwidth,trim={0 0 0 0},clip]{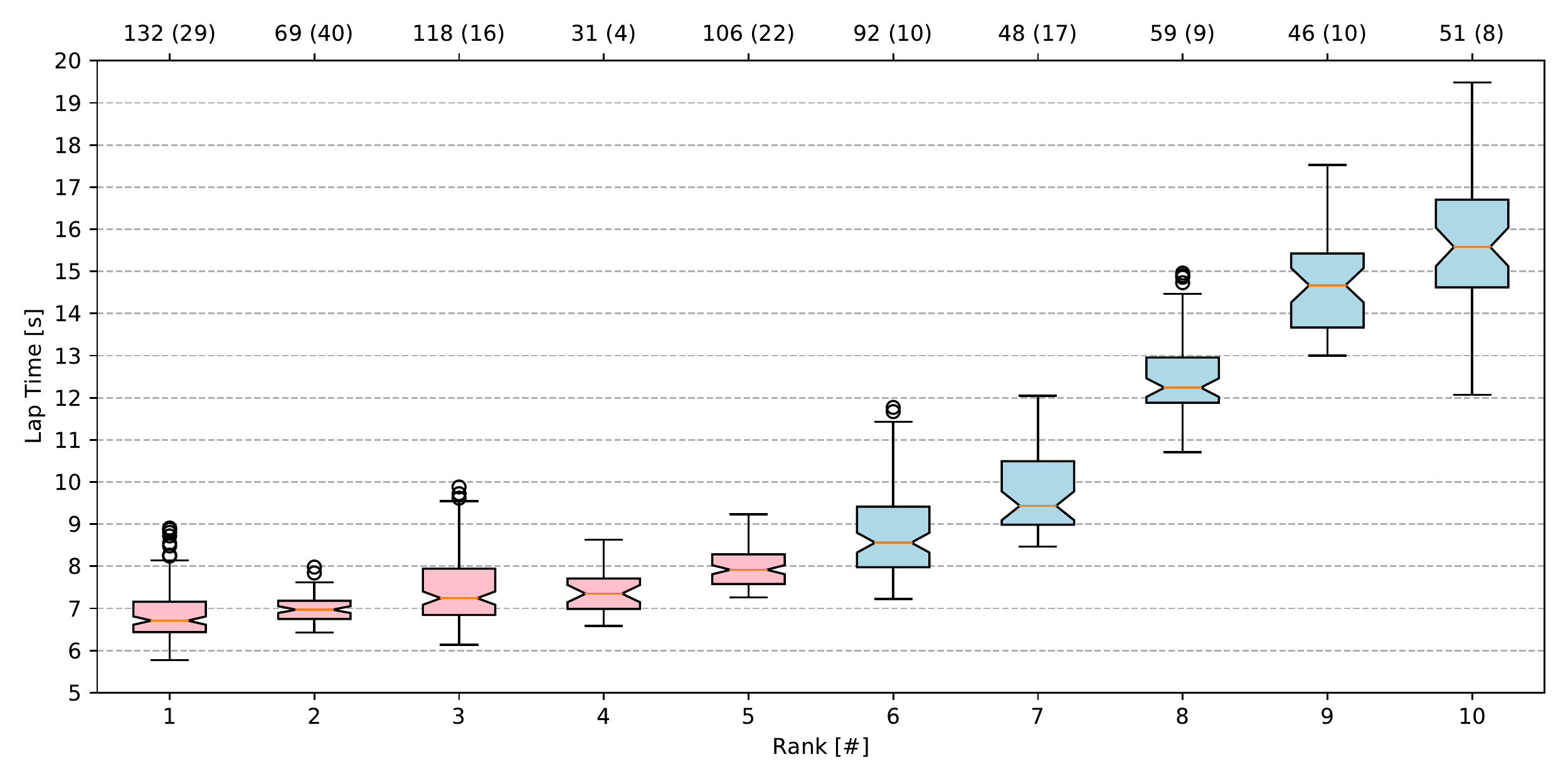}
    \caption[width=\textwidth] {\label{fig:laptimes_by_subject} Overview of lap times of professional (pink) and beginner pilots (blue). Subjects are ranked by median lap time. Values on top indicate the number of completed laps and number of discarded laps (in brackets).}
\end{figure*}

\section{Related Work}
Differences between experts and non-experts are generally seen in ten thousand hours of practice of a skill \cite{Ericsson1993}. However, a meta-analysis showed that practice of a skill accounts only for 18\% of differences between experts and beginners \cite{Macnamara2016}. Sport psychology research shows that cognitive, perceptual abilities, personality traits, and genetics contribute to expert performances. Indeed, differences between experts and novices are found in many sports. For instance, in tennis players, experts show a better ability to predict the trajectory of the tennis ball \cite{Aglioti2008,Rodrigues2002,Savelsbergh2002}. In motorsports, experts as compared to novices generally drive a more optimal race line \cite{VanLeeuwen2017a,Negi2019} and show faster response times \cite{Negi2019,VanLeeuwen2017a}. Some racing drivers also show task-specific adaptations, such as stronger neck stability and grip strength \cite{Backman2005} and more pronounced head movements in curve maneuvers than non-racing drivers. Moreover, racing drivers direct their eye gaze into further distance \cite{Falkmer2005}, use more peripheral vision \cite{Summala1996} and scan the environment more actively as compared to non-racing drivers \cite{Crundall1998,VanLeeuwen2017a}. At present, very few studies have investigated the differences between professional and beginner drone racing pilots. \cite{Barin2017} analyzed video recordings from the 2017 Drone Racing League championships and observed that crashes by professional pilots were mainly related to improper maneuvering, errors in perception, and interference from lighting transitions. These authors had no access to ground truth trajectory data or data from beginner pilots racing the same tracks. \cite{Pfeiffer2021b} studied the flight performance and eye movements of experienced drone racing pilots using a racing simulator and found that eye gaze fixations predicted the direction of the future flight path, were correlated with thrust vector control and showed an average latency of 220 ms between eye gaze and drone thrust vector changes. Whether these effects differ between professional and beginner pilots in a real-world racing scenario remains an open question.

\section{Contributions}
The main contributions of this work are 1. We identify differences between professional and beginner pilots' flight performances and control behavior in a real-world drone racing task on a challenging race track. 2. Using high-resolution drone state position tracking, we identify differences in flight trajectories of professional pilots, beginner pilots, and autonomous drone flying a minimum-time race trajectory. 3. We make our data openly available on the OSF platform (\protect\url{https://osf.io/uabx4/}).

\section{Methods}

\begin{figure*}[]
    \centering
    \includegraphics[width=\textwidth,trim={0 0 0 0},clip]{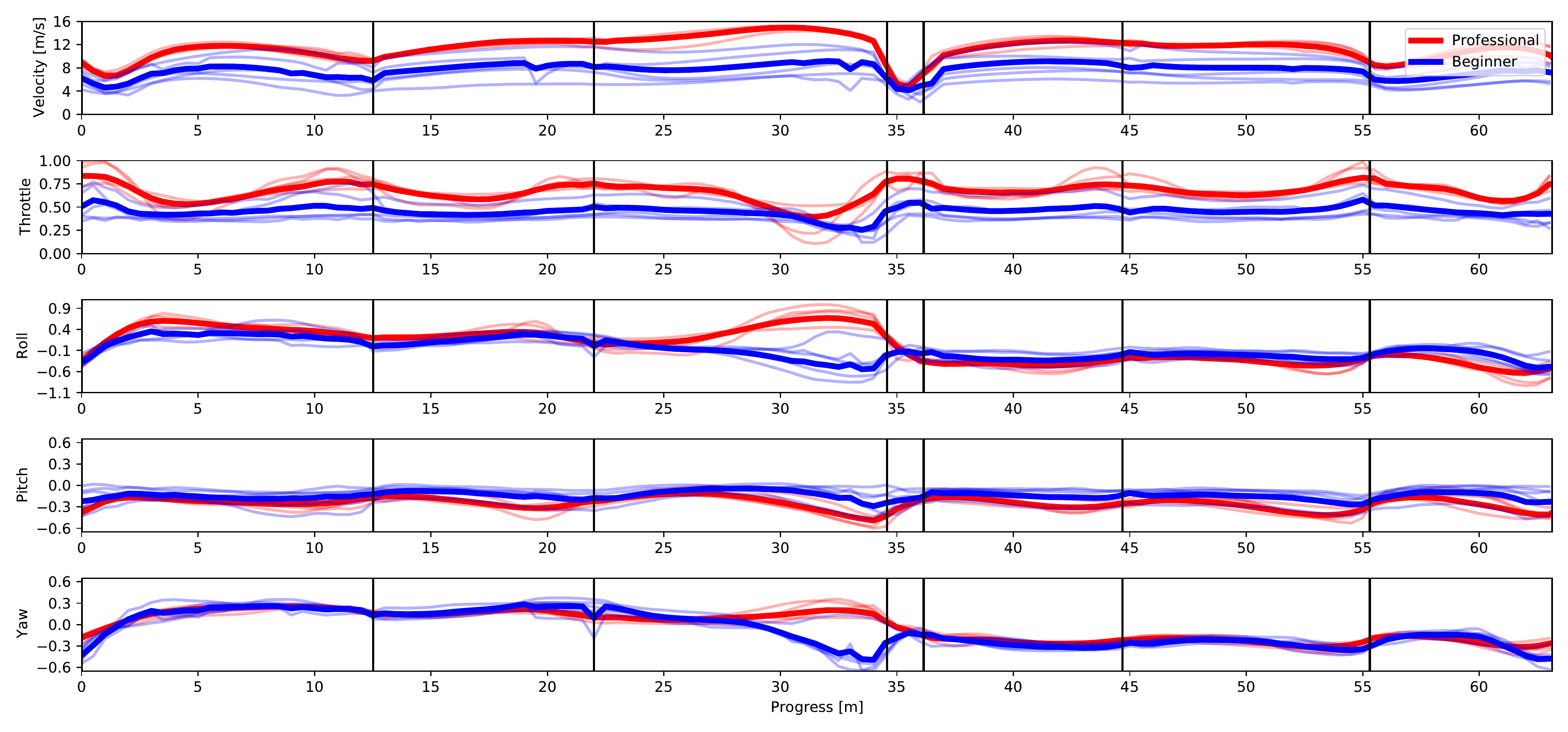}
    \caption[Short caption.] {\label{fig:trpy_timeseries_grandmean} Overview of quadrotor velocity and control inputs (throttle, roll, pitch, and yaw rate) of professional (red) and beginner pilots (blue) as a function of progress along the race track. Subject averages (lighter colors) and group averages (darker colors) are shown. Vertical lines indicate gate passing events.}
\end{figure*}

\subsection{Participants}
Ten experienced drone pilots (all male, all right-handed, average age of 31 years, age range of 15 to 38 years) were recruited online from a local drone racing association and by directly contacting the Swiss National and International drone racing champions. Five pilots who achieved podium places in drone racing championships such as the Drone Racing League, Drone Champions League, and the Swiss Drone League in the 2018-2021 period and who participated in at least ten races are assigned to the professional pilot group. The remaining pilots are assigned to the beginner pilot group. We thus define beginner pilots as pilots with little or no racing experience. Beginner pilots, however, do have prior experience in flying FPV drones for recreational or commercial purposes (e.g., freestyle, aerial photography). According to self-report, all participants declared being n healthy conditions and without a history of neurological or psychological disorder. The study protocol was approved by the ethics committee of the University of Zurich. Participants gave their written informed consent for participating in the study prior to the study. The data recordings were conducted in accordance with the Declaration of Helsinki. Participants received monetary compensation of 30 Swiss Francs per hour for participating in this study.

\subsection{Apparatus and Stimuli}

\textit{Flying arena.} The experiment takes place in the larger Zurich area in Switzerland inside an airplane hangar where a VICON\footnote{\url{https://www.vicon.com/}} system with 36 cameras is installed. This is one of the world’s largest drone flying arenas (30 × 30 × 8 m) that provide the platform with down to millimeter accuracy measurements of position and orientation. A challenging race track is set up in the flying arena consisting of seven gates and three challenging maneuvers (see Fig. \ref{fig:pilot_drone_track}C for an illustration). The race track has a length of 63 m. Drone state ground truth is recorded using VICON Vantage cameras at 200 Hz sampling rate.

\textit{Racing drone}. The racing drone flown by the human pilot (shown in Fig. \ref{fig:pilot_drone_track}B) is built from off-the-shelve components and consists of a carbon fiber frame (Rennstall Stretch-X 5-inch frame), a Hobbywing 60A 4-in-1 Electronic Speed Controller board, a Radix flight controller running Betaflight\footnote{\url{https://betaflight.com/}} firmware (version 4.1), Hobbywing RacePro 2306 1600 kv motors, and Azure Power 5148 propellers. The video system consists of a DJI FPV\footnote{\url{https://www.dji.com}} Airunit camera (120 FPS, 160-degree diagonal field of view) and video transmitter streaming digital video at 25 Mbits/sec at 25 mW (average latency of 25 ms) to the DJI FPV goggles worn by the pilot. Tattu R-Line LiPo batteries (6 cells, 1300 mAh) are used. Five reflective markers for position tracking are attached to the frame using 5 mm carbon fiber sticks. The all-up-weight of the platform is 650g. Drone onboard blackbox logs of control commands IMU, and motor speeds are saved to SD card at 1 kHz. Motor outputs are limited to 63\% in the flight controller software in order to match the maximum thrust-to-weight ratio between human and autonomous drone (i.e., 3.3 thrust-to-weight ratio). 

\textit{Autonomous drone.} The drone used to fly the race track autonomously is an in-house built platform from off-the-shelf components and is identical to the platform used in \cite{Foehn2021,Romero2021}. The autonomous platform consists of an NVIDIA Jetson TX2\footnote{\url{https://developer.nvidia.com/EMBEDDED/jetson-tx2}} board that bridges the control commands via a Laird RM0242\footnote{\url{https://www.lairdconnect.com/wireless-modules/ramp-ism-modules}} wireless
low-latency module. Our method runs in an offboard desktop computer equipped with an Intel(R) Core(TM) i7-8550 CPU @ 1.80GHz. A Radix flight controller board that contains the Betaflight firmware is used as a low-level controller. The all-up weight of the platform is 850g. The method used for autonomous trajectory planning and control is based on model-based optimization using Complementary Process Constraints (CPC) and a Model Predictive Controller (MPC) described in detail in \cite{Foehn2020a,Foehn2021}.

\subsection{Experimental Procedure}
Each pilot is tested individually. After arrival at the flying arena, the pilot is shown the race track and receives the opportunity to walk along the track for closer inspection. Then, the pilot is asked to configure according to personal preference the rate profile in Betaflight, which maps joystick movements to body rate commands and to adjust the FPV camera uptilt angle. The pilot is then allowed to hover the drone to test the settings. Before the experiment, the pilot is instructed to perform a single-player time trials race. The pilot is tasked to fly as fast as possible and as many laps as possible within approximately 2.5 min of flight time per LiPo battery. Pilots fly varying amounts of LiPo batteries depending on their time available (max. 2 hours). 

\subsection{Data Acquisition}
During each flight, the VICON position tracking system records ground truth position and rotation data at 200 Hz, which is saved to rosbags or csv files. In addition, flight controller blackbox logs are saved to an SD card. All gate positions are tracked using VICON markers on the ground floor.

 \begin{figure}[t]
    \centering
    \includegraphics[width=\linewidth,trim={0 0 0 0},clip]{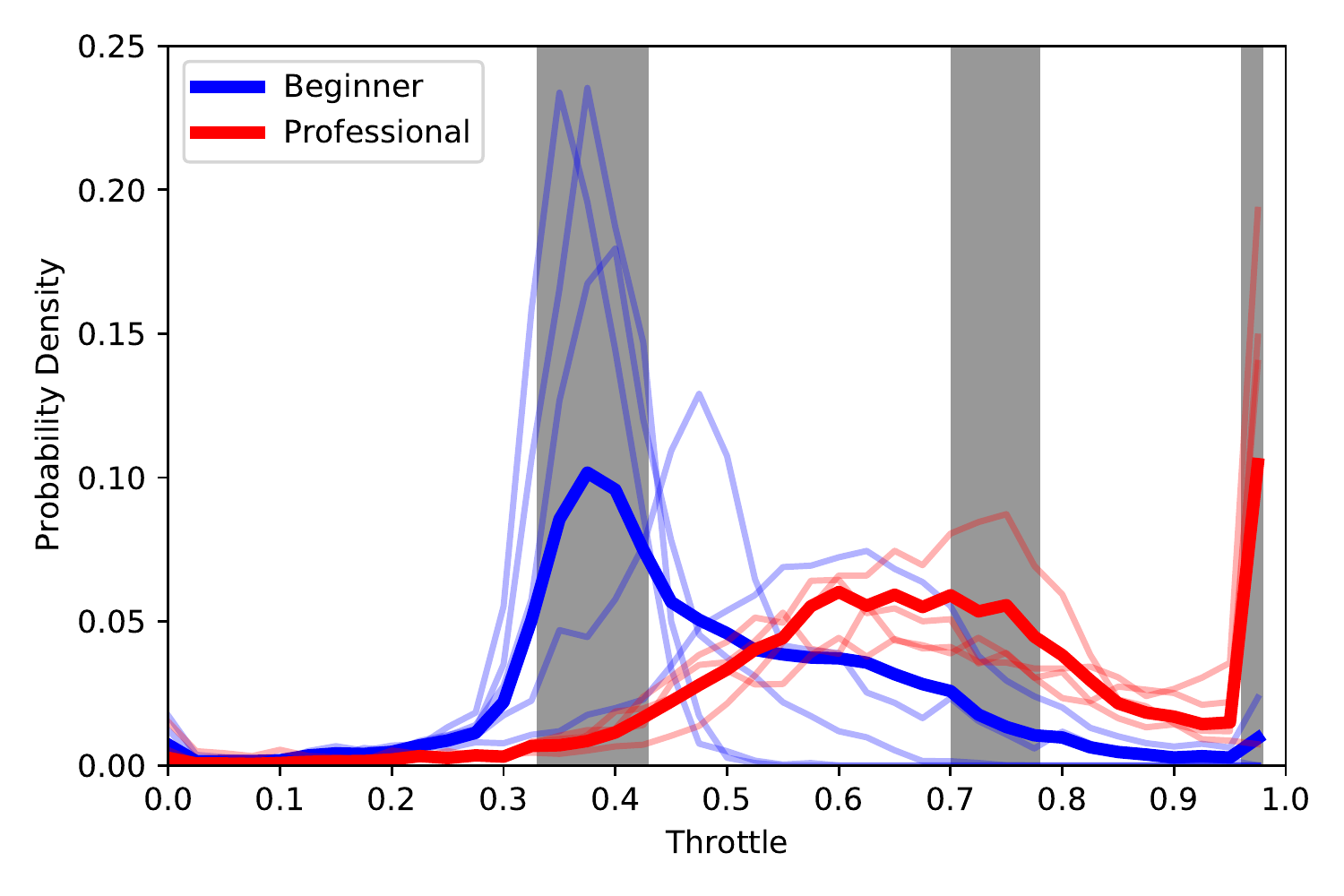}
    \caption[Short caption.] {\label{fig:throttle_histogram_grandmedian} Throttle command usage of beginner pilots (blue) and professional pilots (red) and  showing subject averages (lighter color) and group averages (darker color). Significant differences between professional and beginner pilots are highlighted in gray.}
\end{figure}

\subsection{Data Processing}

\textit{Preprocessing}. The raw blackbox logs are converted from bfl to csv format using the blackbox-decoder software. Drop-out samples in the raw VICON and blackbox data (less than 5\%) are detected and interpolated. Velocity and acceleration data are computed from position data and smoothed using a Hanning filter with 200-ms width). Time-synchronization of blackbox logs and position ground truth data is performed using cross-correlations of norm accelerations derived from position logs and onboard IMU accelerometer data. 

\textit{Segmentation}. After that, seven virtual checkpoints representing the gates are computed from gate position data. Each checkpoint is defined as a plane with 2 x 2 meter dimension positioned at the gate locations. These checkpoints are used to detect gate-passing events from quadrotor position logs. A valid gate pass consists of a signed-distance crossing of the checkpoint plane in the desired flight direction. The gate passing events are subsequently used to identify the start and end of a lap (i.e. passing gate 1) and all intermediate gate passes. After that, we identify valid laps as those laps in which all gates were passed in the desired sequence and invalid laps as those laps where gates were missed or passed in the wrong sequence. Outlier laps are those for which the lap time exceeds two interquartile ranges of the subject median lap time. Invalid laps and outlier laps are discarded from further analysis.

\begin{figure}[t]
    \centering
    \includegraphics[width=\linewidth,trim={0 0 0 0},clip]{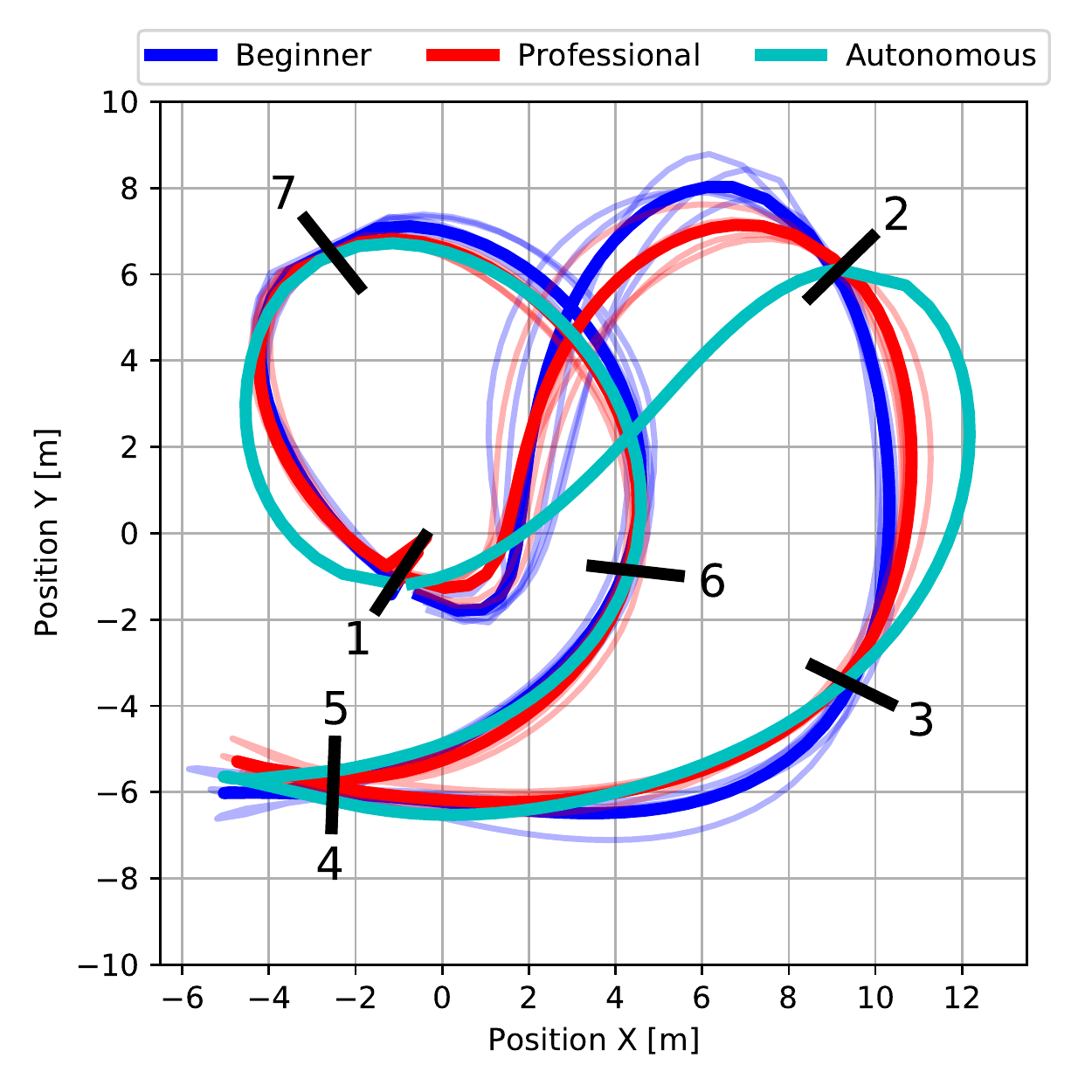}
    \caption[Short caption.] {\label{fig:trajectories_grandmean} Average flight paths of beginner pilots (blue), professional pilots (red), and autonomous drone (cyan). Lighter colors show subject averages and darker colors show the group average. Gates are shown in black and numbered in sequence.}
\end{figure}

\textit{Track progress}. Track progress in meters is computed by projecting the vector pointing from the center of the past gate to the quadrotor position on the vector pointing from the past gate center to the next gate center. The track used in this experiment has maximum track progress of 63 m (see Fig. \ref{fig:trpy_timeseries_grandmean} for an illustration). Data is segmented into 0.5-m progress bins (128 bins).

\textit{Feature extraction}. Flight performance features include the number of attempted, completed, and discarded laps, as well as median and peak velocity. Control command metrics are normalize stick commands (0-1 range) for throttle, roll, pitch, and yaw rate. Trajectory analysis uses the mean squared error between human and autonomous drone positions at each track progress bin.

\subsection{Data Analysis}
 We compare the flight performances, control commands, and flight trajectories of professional and beginner pilots. In addition, we compare pilots’ ability at the professional and beginner level to improve performance on the race track across time. We also perform correlations between flight performance metrics and a pilot’s experience and age. Finally, we compare trajectories by human pilots to minimum-time trajectories flown by the autonomous drone. Statistical analyses are carried out using General Linear Mixed Model analysis. We use a random intercept model with one fixed effect (Group: Professional, Beginner) and one random effect (Subject).

\section{Results}

\subsection{Flight Performance}
Overall, professional pilots attempt 567 laps of which 456 (80\%) are completed and 111 (20\%) are discarded due to the pilot missing a gate or a crash. Beginner pilots attempt 350 laps, of which 296 (84\%) are completed, and 54 (16\%) are discarded. The proportion of the completed out of the attempted laps is slightly smaller for professional pilots as compared to beginner pilots and mainly related to crashes. The majority of crashes take place at gate 2 (60\%), which is a difficult gate to take because it is rotated to at a very shallow angle with respect to the previous gate. The second most frequent crash location is the high gate 4 of the challenging Split-S maneuver (30\% of crashes). The analysis of lap times shows that professional pilots achieved the fastest lap times and fastest median lap times (Fig. \ref{fig:laptimes_by_subject}), indicating that professional pilots fly consistently faster than beginner pilots.

\subsection{Quadrotor Control}
Professional pilots achieve overall higher peak velocity and median velocity as compared to beginner pilots. Fig. \ref{fig:trpy_timeseries_grandmean} shows an overview of the commanded throttle and body rates (roll, pitch, yaw). Two main characteristics can be observed for each pilot group. First, professional pilots use a higher throttle command throughout the race and perform more aggressive turning commands (roll, pitch, and yaw) than beginner pilots. Second, professional pilots initiate the Split-S maneuver earlier and execute it differently than beginner pilots. More specifically, professional pilots perform a rightward flip with a dominant roll component. Beginner pilots, however, execute a leftward rotation with a dominant yaw component. Considering that the Split-S maneuver immediately follows a right turn maneuver, it appears that the rightward flip of professional pilots is more effective at preserving a high forward velocity throughout the maneuver. 

These prominent differences between professional and beginner pilots are further reflected in the distribution of throttle command usage in Fig. \ref{fig:throttle_histogram_grandmedian}. Professional pilots consistently use not only a higher median throttle value (70\%) as compared to beginner pilots (39\%), but also use full throttle (100\%) more frequently than beginner pilots. 

\subsection{Comparison to Minimum-Time Trajectories}
Flight trajectory analysis shows that professional pilots choose a more optimal race line, reflected in a smaller path deviation from the autonomous drone for professional pilots (MSE=1.19 m, SD=0.45 m) as for beginner pilots (MSE=1.89 m, SD=0.43 m; p\textless0.05). Moreover, professional pilots show smaller variance (IQR: M=0.37 m, SD=0.02 m) in their flight paths as compared to beginner pilots (IQR: M=0.46 m, SD=0.06 m, p\textless0.001). Flight paths of human pilots differ significantly from the autonomous drone. Fig. \ref{fig:trajectories_grandmean} shows that these differences are most pronounced at gates 1, 2, and 3, where human pilots choose a gate entry angle that is closer to the gate normal than the autonomous drone. This can be explained by the fact that human pilots must rely on visual feedback from a first-person viewpoint. The pilots appear to optimize the visibility of the gate before passing through it. 

\begin{figure}[t]
    \centering
    \includegraphics[width=\linewidth,trim={0 0 0 0},clip]{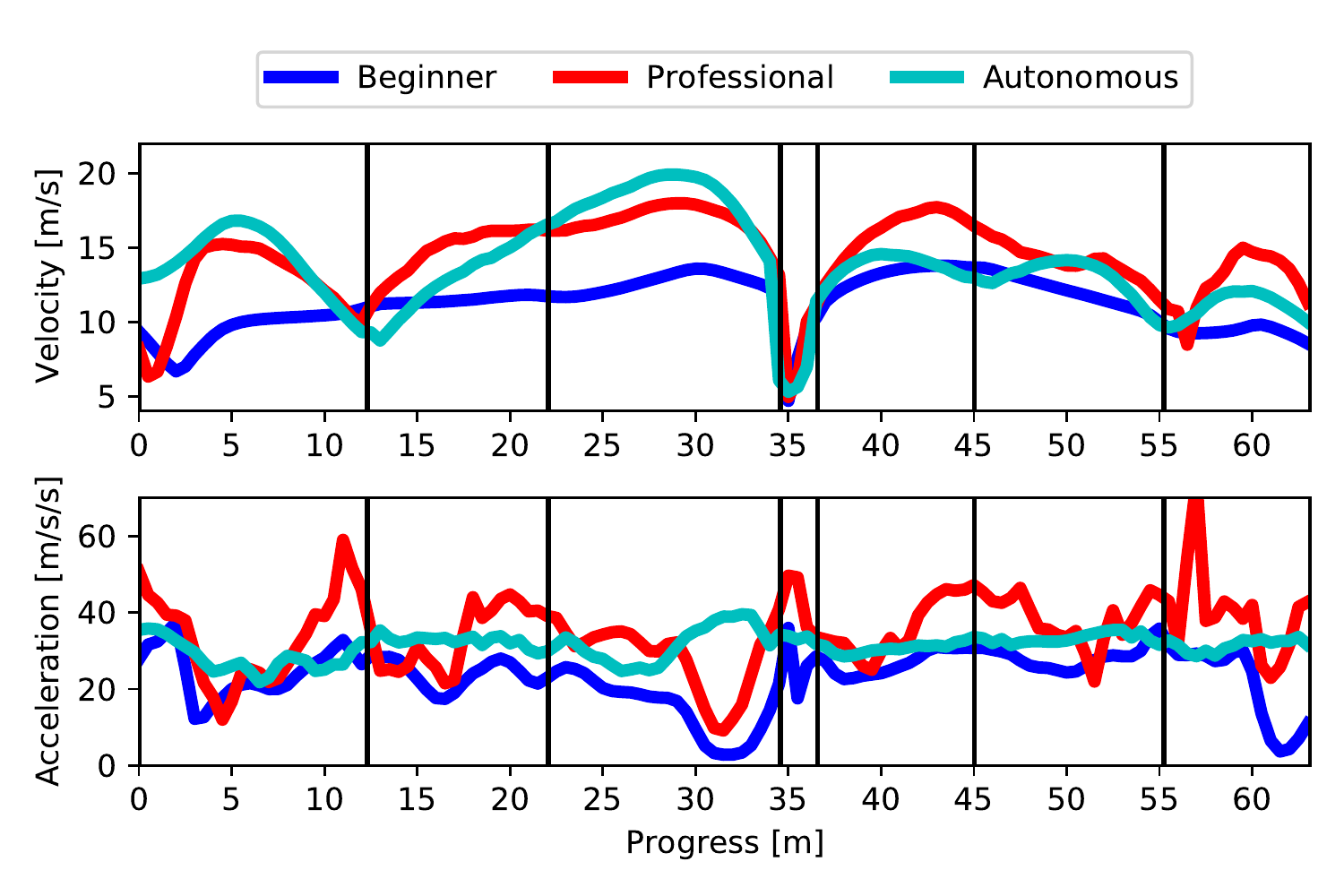}
    \caption[Short caption.] {\label{fig:fastest_lap} Drone velocity and acceleration for fastest lap by beginner pilots (blue), professional pilots (red), and the autonomous drone (cyan) as a function of track progress. Vertical lines show gate passing events.}
\end{figure}

Fig. \ref{fig:fastest_lap} shows a comparison of velocity and acceleration profiles by the fastest human pilot and the autonomous drone flown on the real-world race track. Although the achieved maximum and median velocities are highly similar for human and autonomous flight, a large difference in commanded acceleration can be observed. The human pilot achieves higher acceleration and shows significantly more variance in the commanded accelerations. These results indicate that the human pilot exerts bursts of accelerations that push the platform more to its limits. By contrast, the autonomous drone shows a somewhat lower but more consistent acceleration profile, as in \cite{Foehn2021}. This is related to the thrust-to-weight ratio limits set in the CPC planning algorithm (detailed discussion of this point in \cite{Foehn2021, Romero2021}). More specifically, although the autonomous platform can perform individual maneuvers of up to 4.0 thrust-to-weight ratio, reliable trajectory tracking the MPC is not possible at the platform limits. Therefore \cite{Foehn2021} plan minimum-time trajectories at a 3.3 thrust-to-weight ratio, which allows reliable tracking in the real world. Future work using adaptive control (e.g., Model Predictive Contouring Controller as in \cite{Romero2021}) will enable further to push these limits to the actual platform limits. 

\section{Discussion}
This study investigated the differences between professional and beginner pilots in real-world drone racing. Our results show that professional pilots consistently outperform beginner pilots in terms of lap times, velocity, and aggressiveness of the maneuvers performed (in particular for the Split-S maneuver). Professional pilots can maintain higher thrust throughout the race track and choose more optimal racing lines. Similar differences between racing and non-racing drivers have been found \cite{VanLeeuwen2017a,Negi2019}. The most significant differences in flight trajectories between human pilots and the tested CPC-MPC approach to autonomous flight are that human pilots optimize the visibility of racing gates when choosing an entry angle to a race gate. Human pilots command accelerations closer to the platform limits than the autonomous platform.

What insights can be derived from this study for future work on autonomous fast and agile multi-rotor flight? Our results clearly show a strong contribution of expertise to performance in drone racing. Beginner pilots with little racing experience are unable to achieve the top-level performances of professional pilots. Thus, the observed performances resemble a snapshot of an ongoing learning process of visual perception and motor control skills required for fast and agile flight. In that sense, human pilots resemble reinforcement learning agents who continually explore and learn from past experiences. State of the art model-based approaches to path planning do not consider experiences. Thus, a promising approach for future work is combining learning-based and model-based methods for path planning and control. 

Another important difference between human pilots and the autonomous platform used in this work is that state estimation of the autonomous platform is based on a position tracking system, whereas human pilots perform vision-based state estimation. State of the art vision-based systems for autonomous flight still have not achieved the level of performance of human pilots \cite{Foehn2020}. Thus, future work should work on vision-based perception in challenging conditions with motion blur and limited camera field of view \cite{Delmerico2019}. This work is also an important first step towards deeper insights into human-machine interaction. Besides more technical implications, it is our view that autonomous drone racing is a highly relevant and challenging research topic that will benefit applications in the entertainment and industrial sectors. This technology will further benefit numerous applications, such as infrastructure inspection, search-and-rescue, and autonomous delivery.



\bibliographystyle{IEEEtran}
\bibliography{references}

\begin{thebibliography}{10}
\providecommand{\url}[1]{#1}
\csname url@samestyle\endcsname
\providecommand{\newblock}{\relax}
\providecommand{\bibinfo}[2]{#2}
\providecommand{\BIBentrySTDinterwordspacing}{\spaceskip=0pt\relax}
\providecommand{\BIBentryALTinterwordstretchfactor}{4}
\providecommand{\BIBentryALTinterwordspacing}{\spaceskip=\fontdimen2\font plus
\BIBentryALTinterwordstretchfactor\fontdimen3\font minus
  \fontdimen4\font\relax}
\providecommand{\BIBforeignlanguage}[2]{{%
\expandafter\ifx\csname l@#1\endcsname\relax
\typeout{** WARNING: IEEEtran.bst: No hyphenation pattern has been}%
\typeout{** loaded for the language `#1'. Using the pattern for}%
\typeout{** the default language instead.}%
\else
\language=\csname l@#1\endcsname
\fi
#2}}
\providecommand{\BIBdecl}{\relax}
\BIBdecl

\bibitem{Mellinger2010}
D.~Mellinger, N.~Michael, and V.~Kumar, ``{Trajectory generation and control
  for precise aggressive maneuvers with quadrotors},'' \emph{The International
  Journal of Robotics Research}, vol.~31, no.~5, pp. 664--674, apr 2012.

\bibitem{Loianno2017}
G.~Loianno, C.~Brunner, G.~McGrath, and V.~Kumar, ``{Estimation, Control, and
  Planning for Aggressive Flight with a Small Quadrotor with a Single Camera
  and IMU},'' \emph{IEEE Robotics and Automation Letters}, vol.~2, no.~2, pp.
  404--411, 2017.

\bibitem{Kaufmann2018a}
E.~Kaufmann, M.~Gehrig, P.~Foehn, R.~Ranftl, A.~Dosovitskiy, V.~Koltun, and
  D.~Scaramuzza, ``Beauty and the beast: Optimal methods meet learning for
  drone racing,'' \emph{2019 International Conference on Robotics and
  Automation (ICRA)}, pp. 690--696, 2019.

\bibitem{Foehn2020}
P.~Foehn, D.~Brescianini, E.~Kaufmann, T.~Cieslewski, M.~Gehrig, M.~Muglikar,
  and D.~Scaramuzza, ``Alphapilot: Autonomous drone racing,'' \emph{Robotics:
  Science and Systems}, 2020.

\bibitem{Romero2021}
A.~Romero, S.~Sun, P.~Foehn, and D.~Scaramuzza, ``{Model predictive contouring
  control},'' \emph{arXiv preprint arXiv:2108.13205}, 2021.

\bibitem{Moon2019}
H.~Moon, J.~Martinez-Carranza, T.~Cieslewski, M.~Faessler, D.~Falanga,
  A.~Simovic, D.~Scaramuzza, S.~Li, M.~Ozo, C.~{De Wagter}, G.~de~Croon,
  S.~Hwang, S.~Jung, H.~Shim, H.~Kim, M.~Park, T.-C. Au, and S.~J. Kim,
  ``{Challenges and implemented technologies used in autonomous drone
  racing},'' \emph{Intelligent Service Robotics}, vol.~12, no.~2, pp. 137--148,
  apr 2019.

\bibitem{Cocoma-Ortega2019}
J.~A. Cocoma-Ortega and J.~Mart{\'{i}}nez-Carranza, ``{Towards High-Speed
  Localisation for Autonomous Drone Racing},'' in \emph{Advances in Soft
  Computing}, L.~Mart{\'{i}}nez-Villase{\~{n}}or, I.~Batyrshin, and
  A.~Mar{\'{i}}n-Hern{\'{a}}ndez, Eds.\hskip 1em plus 0.5em minus 0.4em\relax
  Cham: Springer International Publishing, 2019, pp. 740--751.

\bibitem{Madaan2020a}
R.~Madaan, N.~Gyde, S.~Vemprala, M.~Brown, K.~Nagami, T.~Taubner,
  E.~Cristofalo, D.~Scaramuzza, M.~Schwager, and A.~Kapoor, ``Airsim drone
  racing lab,'' \emph{PMLR post-proceedings of the NeurIPS 2019's Competition
  Track}, 2020.

\bibitem{Guerra2019}
W.~Guerra, E.~Tal, V.~Murali, G.~Ryou, and S.~Karaman, ``{FlightGoggles:
  Photorealistic Sensor Simulation for Perception-driven Robotics using
  Photogrammetry and Virtual Reality},'' in \emph{2019 IEEE/RSJ International
  Conference on Intelligent Robots and Systems (IROS)}.\hskip 1em plus 0.5em
  minus 0.4em\relax IEEE, nov 2019, pp. 6941--6948.

\bibitem{cordisAgileFlightEUproject}
\BIBentryALTinterwordspacing
``{CORDIS - European Comission. AgileFlight.}'' 2021. [Online]. Available:
  \url{https://cordis.europa.eu/project/id/864042}
\BIBentrySTDinterwordspacing

\bibitem{Mohta2018}
K.~Mohta, M.~Watterson, Y.~Mulgaonkar, S.~Liu, C.~Qu, A.~Makineni, K.~Saulnier,
  K.~Sun, A.~Zhu, J.~Delmerico, K.~Karydis, N.~Atanasov, G.~Loianno,
  D.~Scaramuzza, K.~Daniilidis, C.~J. Taylor, and V.~Kumar, ``{Fast, autonomous
  flight in GPS-denied and cluttered environments},'' \emph{Journal of Field
  Robotics}, vol.~35, no.~1, pp. 101--120, jan 2018.

\bibitem{Pfeiffer2021b}
C.~Pfeiffer and D.~Scaramuzza, ``{Human-piloted drone racing: Visual processing
  and control},'' \emph{IEEE Robotics and Automation Letters}, vol.~6, no.~2,
  pp. 3467--3474, 2021.

\bibitem{Barin2017}
A.~Barin, I.~Dolgov, and Z.~O. Toups, ``{Understanding dangerous play: A
  grounded theory analysis of high-performance drone racing crashes},'' in
  \emph{CHI PLAY 2017 - Proceedings of the Annual Symposium on Computer-Human
  Interaction in Play}, 2017, pp. 485--496.

\bibitem{Ericsson1993}
K.~A. Ericsson, R.~T. Krampe, and C.~Tesch-R{\"{o}}mer, ``{The role of
  deliberate practice in the acquisition of expert performance.}''
  \emph{Psychological Review}, vol. 100, no.~3, pp. 363--406, 1993.

\bibitem{Macnamara2016}
B.~N. Macnamara, D.~Moreau, and D.~Z. Hambrick, ``{The Relationship Between
  Deliberate Practice and Performance in Sports},'' \emph{Perspectives on
  Psychological Science}, vol.~11, no.~3, pp. 333--350, may 2016.

\bibitem{Aglioti2008}
S.~M. Aglioti, P.~Cesari, M.~Romani, and C.~Urgesi, ``{Action anticipation and
  motor resonance in elite basketball players},'' \emph{Nature Neuroscience},
  vol.~11, no.~9, pp. 1109--1116, sep 2008.

\bibitem{Rodrigues2002}
S.~T. Rodrigues, J.~N. Vickers, and A.~M. Williams, ``{Head, eye and arm
  coordination in table tennis},'' \emph{Journal of Sports Sciences}, vol.~20,
  no.~3, pp. 187--200, jan 2002.

\bibitem{Savelsbergh2002}
G.~J. Savelsbergh, A.~M. Williams, J.~V.~D. Kamp, and P.~Ward, ``{Visual
  search, anticipation and expertise in soccer goalkeepers},'' \emph{Journal of
  Sports Sciences}, vol.~20, no.~3, pp. 279--287, jan 2002.

\bibitem{VanLeeuwen2017a}
P.~M. van Leeuwen, S.~de~Groot, R.~Happee, and J.~C.~F. de~Winter,
  ``{Differences between racing and non-racing drivers: A simulator study using
  eye-tracking},'' \emph{PLOS ONE}, vol.~12, no.~11, p. e0186871, nov 2017.

\bibitem{Negi2019}
N.~S. Negi, P.~{Van Leeuwen}, and R.~Happee, ``{Differences in driver behaviour
  between race and experienced drivers: A driving simulator study},''
  \emph{VEHITS 2019 - Proceedings of the 5th International Conference on
  Vehicle Technology and Intelligent Transport Systems}, pp. 360--367, 2019.

\bibitem{Backman2005}
J.~Backman, K.~H{\"{a}}kkinen, J.~Ylinen, A.~H{\"{a}}kkinen, and
  H.~Kyr{\"{o}}l{\"{a}}inen, ``{Neuromuscular Performance Characteristics of
  Open-Wheel and Rally Drivers},'' \emph{The Journal of Strength and
  Conditioning Research}, vol.~19, no.~4, p. 777, 2005.

\bibitem{Falkmer2005}
T.~Falkmer and N.~P. Gregersen, ``{A comparison of eye movement behavior of
  inexperienced and experienced drivers in real traffic environments},''
  \emph{Optometry and Vision Science}, vol.~82, no.~8, pp. 732--739, 2005.

\bibitem{Summala1996}
H.~Summala, T.~Nieminen, and M.~Punto, ``{Maintaining Lane Position with
  Peripheral Vision during In-Vehicle Tasks},'' \emph{Human Factors: The
  Journal of the Human Factors and Ergonomics Society}, vol.~38, no.~3, pp.
  442--451, sep 1996.

\bibitem{Crundall1998}
D.~E. Crundall and G.~Underwood, ``{Effects of experience and processing
  demands on visual information acquisition in drivers},'' \emph{Ergonomics},
  vol.~41, no.~4, pp. 448--458, apr 1998.

\bibitem{Foehn2021}
P.~Foehn, A.~Romero, and D.~Scaramuzza, ``{Time-optimal planning for quadrotor
  waypoint flight},'' \emph{Science Robotics}, vol.~6, no.~56, jul 2021.

\bibitem{Foehn2020a}
------, ``Time-optimal planning for quadrotor waypoint flight,'' \emph{Science
  Robotics}, vol.~6, no.~56, 2021.

\bibitem{Delmerico2019}
J.~Delmerico, T.~Cieslewski, H.~Rebecq, M.~Faessler, and D.~Scaramuzza, ``{Are
  We Ready for Autonomous Drone Racing? The UZH-FPV Drone Racing Dataset},'' in
  \emph{2019 International Conference on Robotics and Automation (ICRA)}, vol.
  2019-May.\hskip 1em plus 0.5em minus 0.4em\relax IEEE, may 2019, pp.
  6713--6719.

\end{thebibliography}

\end{document}